\documentclass[lettersize,journal]{IEEEtran}
\usepackage{amsmath,amsfonts}
\usepackage{algorithmic}
\usepackage{algorithm}
\usepackage{array}
\usepackage{textcomp}
\usepackage{stfloats}
\usepackage{subfigure}
\usepackage{url}
\usepackage{verbatim}
\usepackage{xcolor}
\usepackage{graphicx}
\usepackage{cite}
\begin{document}

\title{AI$^2$MMUM: AI-AI Oriented Multi-Modal Universal Model Leveraging Telecom Domain Large Model}

\author{Tianyu Jiao, Zhuoran Xiao, Yihang Huang, Chenhui Ye, Yijia Feng, Liyu Cai, Jiang Chang,\\Fangkun Liu, Yin Xu, Dazhi He, Yunfeng Guan, and Wenjun Zhang,~\IEEEmembership{Fellow,~IEEE}
\thanks{Manuscript received xxx; revised xxx; accepted xxx. This work was supported in part by the National Key Research and Development Program of China under Grant 2024YFE0200600.}
\thanks{Tianyu Jiao (\textit{Student Member, IEEE}) is with the Cooperative Medianet Innovation Center, Shanghai Jiao Tong University, Shanghai 200240, China, and also with the Nokia Bell Labs, Shanghai 201206, China (e-mail: jiaotianyu@sjtu.edu.cn).}
\thanks{Zhuoran Xiao (\textit{Member, IEEE}), Chenhui Ye (\textit{Member, IEEE}), Yijia Feng (\textit{Member, IEEE}), Liyu Cai, Jiang Chang, and Fangkun Liu are with the Nokia Bell Labs, Shanghai 201206, China (e-mail: \{zhuoran.xiao, chenhui.a.ye, yijia.feng, liyu.cai, jiang.chang\}@nokia-sbell.com, frankenl@usc.edu).}
\thanks{Yin Xu (\textit{Senior Member, IEEE}), Dazhi He (\textit{Senior Member, IEEE}), Yunfeng Guan, and Wenjun Zhang (\textit{Fellow, IEEE}) are with the Cooperative Medianet Innovation Center, Shanghai Jiao Tong University, Shanghai 200240, China (e-mail: \{xuyin, hedazhi, yfguan69, zhangwenjun\}@sjtu.edu.cn).}
\thanks{Yihang Huang (\textit{Member, IEEE}) is with School of Communications and Information Engineering and also with Institute of Intelligent Communications and Network Security, Chongqing University of Posts and Telecommunications, Chongqing 400065, China (e-mail: huangyihang@cqupt.edu.cn).}
\thanks{The corresponding author is Yin Xu.}}


\maketitle

\begin{abstract}
Designing a 6G-oriented universal model capable of processing multi-modal data and executing diverse air interface tasks has emerged as a common goal in future wireless systems. Building on our prior work in communication multi-modal alignment and telecom large language model (LLM), we propose a scalable, task-aware artificial intelligence-air interface multi-modal universal model (AI$^2$MMUM), which flexibility and effectively perform various physical layer tasks according to subtle task instructions. The LLM backbone provides robust contextual comprehension and generalization capabilities, while a fine-tuning approach is adopted to incorporate domain-specific knowledge. To enhance task adaptability, task instructions consist of fixed task keywords and learnable, implicit prefix prompts. Frozen radio modality encoders extract universal representations and adapter layers subsequently bridge radio and language modalities. Moreover, lightweight task-specific heads are designed to directly output task objectives. Comprehensive evaluations demonstrate that AI$^2$MMUM achieves SOTA performance across five representative physical environment/wireless channel-based downstream tasks using the WAIR-D and DeepMIMO datasets.
\end{abstract}


\vspace{-0.2cm}
\section{Introduction}
\begin{figure*}[t]
\centering
\includegraphics[width=2\columnwidth]{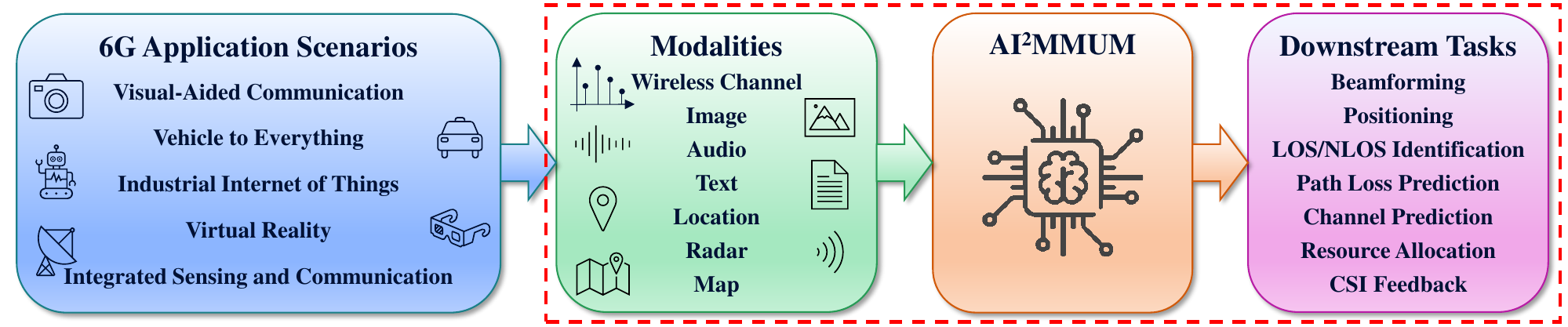}
\caption{The 6G-oriented AI$^2$MMUM capable of processing wireless multi-modal data and performing various air interface tasks.}
\label{f1}
\vspace{-0.2cm}
\end{figure*}

The vision for 6G networks is to enable pervasive intelligence with native support for artificial intelligence (AI). However, traditional wireless AI models are typically designed for specific tasks with minimal parameters, leading to limited transferability and exponentially increased system complexity and model management challenges, which are unsustainable. A universal model is envisioned to perform diverse tasks with high accuracy by leveraging extensive parameters, massive data, and significant computational resources for knowledge integration, reasoning, and generalization \cite{r0}. In the 6G era, emerging technologies such as integrated sensing and communication (ISAC), vision-aided communication, and vehicle to everything (V2X) will greatly enrich wireless modalities. This evolution necessitates developing an AI-air interface multi-modal universal model (AI$^2$MMUM) capable of handling diverse data--such as vision, maps, location, wireless channels, and radar--and executing multiple tasks, as depicted in Fig. \ref{f1}.

Currently, there are only preliminary visions and limited attempts for AI$^2$MMUM, leaving many open questions. For instance, using uplink channel and 3D environment data, the universal model is expected to perform downlink beamforming and power allocation \cite{r1}. \cite{r2} proposed a wireless-centric foundation model, integrating capabilities such as multi-modal data fusion, grounding, and instructibility. In wireless-related fields, 2D images, 3D LiDAR point clouds, and map contexts were incorporated into large language models (LLMs) for map and traffic scene understanding in V2X scenarios \cite{r3}. Cross-modal fusion of mmWave radar and natural language enables 3D visual grounding, facilitating environmental understanding in autonomous driving \cite{r4}. Additionally, NetLLM was developed to allow LLMs to process multi-modal networking data and generate task-specific answers \cite{r5}. However, the development of wireless AI$^2$MMUM still lacks systematic model structure designs and cost-effective, flexible training methods.

To develop such an AI$^2$MMUM, several key challenges must be addressed. Firstly, it should be able to understand task instructions, thereby facilitating accurate task fulfillment. Secondly, the network modules processing wireless multi-modal data must exhibit robust feature extraction capability to ensure information integrity. The gap between distinct modalities should also be effectively bridged to enable seamless knowledge fusion. Moreover, to execute multiple tasks within a single model, the AI$^2$MMUM backbone must possess strong multi-modal context comprehension and generalization capabilities. Finally, the model should be capable of generating task objectives with varying structural and precision demands.

In pursuit of this vision, we propose a scalable, task-aware AI$^2$MMUM. Fixed task keywords and learnable prefix prompts enhance model expressiveness while maintaining semantic consistency. Modality encoders from our prior work on wireless multi-modal alignment extract rich, task-agnostic representations, with adapter modules bridging wireless and language modalities \cite{r6} \cite{r6.5}. A telecom LLM backbone offers integration capability, and low-rank adaptation (LoRA) incorporates domain-specific knowledge while preserving original language knowledge. Lightweight task-specific heads directly produce final objectives. Experimental results demonstrate that our method effectively extracts task-related features based on instructions and outperforms traditional non-LLM methods and models lacking our innovations across multiple sub-tasks.

\vspace{-0.2cm}
\section{Channel Model and Problem Statement}
Taking the wireless channel modality in multi-modal data as an example, we consider a massive multiple-input multiple-output (MIMO) system operating in orthogonal frequency division multiplexing (OFDM) mode with $N_c$ subcarriers. The base station (BS) is equipped with $N_t$ antennas arranged in a uniform linear array (ULA), while the user equipment (UE) has a single antenna. Consequently, the wireless channel between the BS and the UE can be written as,

\begin{equation}
\label{eq1}
\mathrm{\mathbf{h}}(f)=\sum_{i=1}^{N_{\mathrm{path}}}\alpha_{i} e^{-j2\pi f\tau_i} \mathrm{\mathbf{a}}(\theta_i),
\end{equation}
where $f$ is the carrier frequency, $N_{\mathrm{path}}$ denotes the number of propagation paths, $\alpha_i$ represents the amplitude attenuation, $\tau_i$ is the time delay, and $\theta_i$ is the angle of arrival (AoA) of the $i$-th path. Moreover, $\mathrm{\mathbf{a}}(\theta_i)$ is the array vector expressed as,
\begin{equation}
\label{eq2}
\mathrm{\mathbf{a}}(\theta_i)=[1,e^{-j\beta\cos{\theta_i}},\cdots,e^{-j\beta(N_t-1)\cos{\theta_i}}]^T,
\end{equation}
where $\beta=2\pi df/c$, $d$ is the antenna spacing, and $c$ is the speed of light. Consequently, the channel state information (CSI) matrix $\mathrm{\mathbf{H}} \in \mathbb{C}^{N_t \times N_c}$ can be defined as,
\begin{equation}
\label{eq3}
\mathrm{\mathbf{H}}=[\mathrm{\mathbf{h}}(f_1),\mathrm{\mathbf{h}}(f_2),\cdots,\mathrm{\mathbf{h}}(f_{N_c})],
\end{equation}
where $\{f_i \mid i=1,2,\cdots,N_c\}$ is the set of subcarrier frequencies.

Leveraging the strong generalization capability of AI$^2$MMUM, the model is expected to perform diverse air interface tasks tailored to varying requirements using massive collected wireless data. Taking CSI as a representative case, it serves as a mapping of the physical environment in the signal space, supporting a range of air interface tasks such as positioning, line-of-sight (LOS)/non-line-of-sight (NLOS) identification, and mmMIMO precoding, each focusing on distinct features. To address this, the wireless channel $\mathrm{\mathbf{H}}$ and task instruction $\mathrm{L}$ are first transformed into token embedding vectors $\mathrm{\mathbf{E_H}}$ and $\mathrm{\mathbf{E_L}}$, respectively:
\begin{equation}
\label{eq4}
\mathrm{\mathbf{E_H}}=f_{\mathrm{\mathbf{H}}}(\mathrm{\mathbf{H}};\Theta_{\mathrm{\mathbf{H}}}),
\end{equation}
\begin{equation}
\label{eq5}
\mathrm{\mathbf{E_L}}=f_{\mathrm{L}}(\mathrm{L};\Theta_{\mathrm{L}}),
\end{equation}
where $\Theta_{\mathrm{\mathbf{H}}}$ and $\Theta_{\mathrm{L}}$ represent the respective neural network (NN) parameters. To enable cross-modal understanding, $\mathrm{\mathbf{E_H}}$ and $\mathrm{\mathbf{E_L}}$ are designed to share the same dimensionality. Then, these embeddings are concatenated and fed into the AI$^2$MMUM backbone for task-specific feature extraction:
\begin{equation}
\label{eq6}
\mathrm{\mathbf{E_B}}=f_{\mathrm{B}}(\mathrm{Concat(\mathbf{E_H},\mathbf{E_L})};\Theta_{\mathrm{B}}),
\end{equation}
where $\Theta_{\mathrm{B}}$ denotes the backbone parameters. Finally, the feature $\mathrm{\mathbf{E_B}}$ is transformed into the sub-task objective $\mathrm{T}$:
\begin{equation}
\label{eq7}
\mathrm{T}=f_{\mathrm{T}}(\mathrm{\mathbf{E_B}};\Theta_{\mathrm{T}}),
\end{equation}
with $\Theta_{\mathrm{T}}$ representing its NN parameters.

\begin{figure*}[t]
\centering
\includegraphics[width=2\columnwidth]{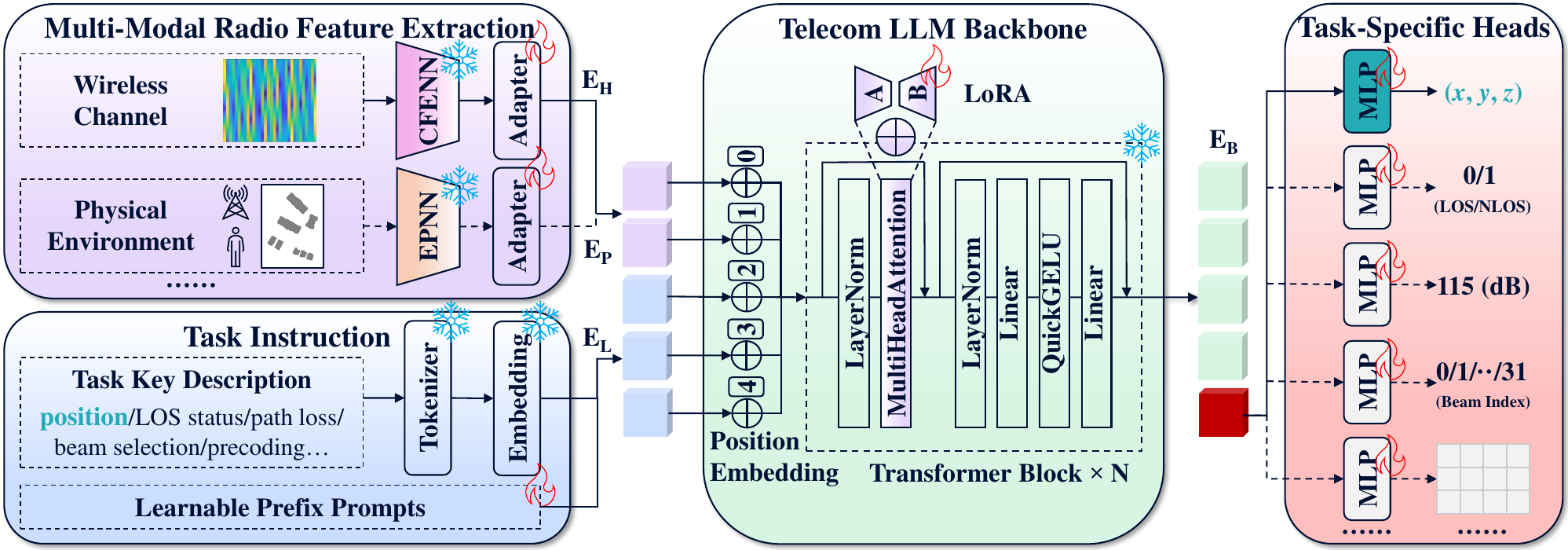}
\caption{Network structure of the proposed 6G-oriented, scalable, and task-aware AI$^2$MMUM.}
\label{f2}
\vspace{-0.3cm}
\end{figure*}

\section{Proposed Network Structure}
In this section, we propose a 6G-oriented, scalable, and task-aware AI$^2$MMUM consisting of four key components: a multi-modal radio feature extraction module, a task instruction module, a telecom LLM backbone enhanced with LoRA, and task-specific heads, as shown in Fig. \ref{f2}.
\vspace{-0.3cm}
\subsection{Multi-Modal Radio Feature Extraction Module}
\begin{figure}[t]
\centering
\includegraphics[width=0.85\columnwidth]{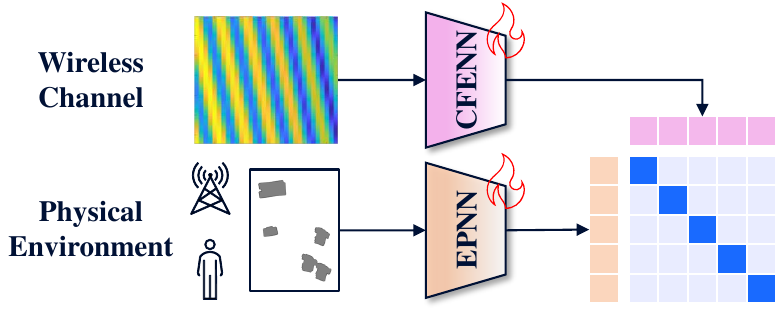}
\caption{Framework for communication multi-modal alignment.}
\label{f3}
\vspace{-0.3cm}
\end{figure}

A fundamental prerequisite for establishing AI$^2$MMUM is the development of robust multi-modal radio encoders capable of extracting informative features from wireless data. In our previous work, we pre-trained a large-scale model comprising an environment perception neural network (EPNN) and a channel feature extraction neural network (CFENN) on extensive datasets encompassing physical environment modality (area, BS, and UE information) and wireless channel modality (CSI), as illustrated in Fig. \ref{f3} \cite{r6.5}. By leveraging contrastive learning, the feature similarities of related environment-channel pairs were maximized, while those of unrelated pairs were minimized. The pre-trained EPNN and CFENN exhibit scenario generalization capabilities and can extract universal modality representations, which are frozen and subsequently applied in this study. Compared to training similar models from scratch using limited local data, these encoders provide more comprehensive insights into wireless characteristics, significantly reducing reliance on labeled data and enhancing AI$^2$MMUM's generalization and task adaptability.

To address the dimensional mismatch between radio modality encoder outputs and LLM backbone inputs, adapter layers are employed to bridge the embedding spaces of radio and language modalities, facilitating seamless cross-modal knowledge transfer and integration, as shown in Fig. \ref{f2}. These layers contain few parameters, incurring minimal computational overhead. When new modalities such as radar and LiDAR are introduced, only adapter layers need to be updated while the corresponding modality encoders remain frozen.

\vspace{-0.3cm}
\subsection{Task Instruction Module}
Task instructions steer AI$^2$MMUM in processing wireless data and executing specific tasks by providing discriminative information within a multi-modal context. Language-based instructions are human-friendly and LLM-compatible. The text is first tokenized into vocabulary indices and then mapped to high-dimensional token embeddings. Task-specific prompts can vary widely. For instance, positioning prompts might be “Please infer the user's position from this CSI.” or “What is the position information of this wireless channel?” Given the unpre-trained wireless data and the black-box characteristics of LLM, the impact of such prompts on task performance remains uncertain. However, consistent task-specific keywords, such as “position”, exist across these prompts. Thus, we propose constructing optimal task-specific prompts by integrating fixed task keyword embeddings with learnable prefix prompts. The latter are trainable embeddings that implicitly encode task instructions and consist of multiple tokens. This design enhances AI$^2$MMUM's transferability and aligns with human cognition, providing a robust solution for diverse air interface tasks.

\vspace{-0.3cm}
\subsection{Telecom LLM Backbone Enhanced with LoRA}
To process and interpret both wireless data and task instructions, their token embeddings $\mathrm{\mathbf{E_H}}$ ($\mathrm{\mathbf{E_P}}$) and $\mathrm{\mathbf{E_L}}$ are concatenated and fed into the LLM backbone. Positional embeddings are added to the combined tokens, providing sequential information for accurate multi-modal context handling. Besides, the backbone comprises multiple stacked transformer blocks that progressively extract and integrate features from language and wireless data through self-attention mechanisms and feedforward networks. This layered structure allows the model to assimilate complex knowledge, thereby enhancing its expressive capacity for sophisticated representation.

However, the LLM backbone, pre-trained with large-scale parameters on extensive natural language corpora, is costly to update and inherently limited in handling wireless multi-modal data such as CSI. Moreover, it must retain its original language knowledge while absorbing communication domain knowledge. To address this, we adopt a flexible and cost-effective fine-tuning method named LoRA. By dynamically adapting specific modules in the backbone, LoRA improves transferability and performance, particularly for small datasets, cross-domain tasks, and resource-constrained scenarios. For each pre-trained matrix $W_0$ of dimension $a \times b$, LoRA assumes a low rank $r \ll \min\{a,b\}$ and constructs two low-rank matrices $A \in \mathbb{R}^{a \times r}$ and $B \in \mathbb{R}^{r \times b}$ to approximate model weight updates as $W=W_0+AB$, significantly reducing the number of tuning parameters, as illustrated in Fig. \ref{f2}. During adaptation, $W_0$ remains frozen, and parameter updates are confined to $A$ and $B$. Due to the unique intrinsic properties of each modality, a LoRA specializes in learning an individual wireless modality, while task instructions and LLM backbone support the completion of diverse downstream tasks related to that modality. Multiple LoRAs can share the same backbone, enabling rapid modality switching and high scalability.

\vspace{-0.5cm}
\subsection{Task-Specific Heads}
To handle tokens of varying lengths, LLMs typically use the last token of the output sequence from a single forward pass as the predicted result for that iteration. Performing multiple iterations followed by de-embedding and de-tokenizing several predicted tokens to obtain language-based results would reduce prediction accuracy and increase computational costs. A single-pass approach can focus the backbone solely on the original multi-modal input, minimizing output uncertainty, reducing inference time, and encapsulating task-related summary features within a single predicted token. To transform the predicted token embedding into downstream task objectives, task-specific heads comprising a single linear layer are employed, enhancing task processing within the LLM backbone and simplifying the external network structure. In practical applications, these heads can be encapsulated within diverse application programming interfaces (APIs). The AI$^2$MMUM can identify and invoke the appropriate API based on task instructions and function calls, producing the final results.

\vspace{-0.2cm}
\section{Experimental Results and Analysis}
\subsection{Dataset Generation and Model Settings}
The wireless AI research dataset (WAIR-D) \cite{r7} and the DeepMIMO dataset \cite{r7.5} are utilized in this experiment. WAIR-D comprises 10,000 real-world areas of varying sizes. In DeepMIMO, we employ the Outdoor 1 (O1) scenario, which features 18 BSs and UEs positioned in a cross-shaped area surrounded by buildings. Our prior work on communication multi-modal alignment utilized 2.25$\mathrm{M}$ modality sample pairs, encompassing physical environment data (area maps, BS positions, and UE positions) and wireless channel data (CSIs), from 9,000 WAIR-D areas numbered \#01001 to \#10000, resulting in the EPNN and CFENN models with robust scenario generalization and modality representation capabilities \cite{r6} \cite{r6.5}. In this study, we train and test the proposed AI$^2$MMUM using 10,000 samples from each of the previously unseen WAIR-D areas \#00032 and \#00247, as well as DeepMIMO O1 BS\#12.

The detailed network structure of the proposed AI$^2$MMUM is depicted in Fig. \ref{f2}. The EPNN and CFENN contain approximately 7.5$\mathrm{M}$ and 7.1$\mathrm{M}$ parameters, respectively, and generate universal modality representations with a dimensionality of 128. The task key description comprises at most two tokens, while the learnable prefix prompts occupy three tokens. The tokenizer layer, embedding layer, and LLM backbone are derived from our telecom LLM, which is retrained from the LLaMA2-7B model using a telecommunication corpus, featuring a token embedding dimension of 4096 \cite{r8}. To align dimensionality, the adapter module employs a linear layer with dimensions from 128 to 4096. To enable the LLM backbone to efficiently acquire environment and channel knowledge, the query and key weight matrices in the self-attention mechanism are fine-tuned using LoRA matrices with a rank of 8, adding approximately 8.4$\mathrm{M}$ parameters. Finally, task-specific heads utilize a linear layer to transform the 4096-dimensional task-related feature token into the downstream task objective.

\begin{table}[t]
\setlength{\tabcolsep}{2.5pt}
\caption{Task Types, Inputs, and Outputs of the Five Exemplary Downstream Tasks}
\vspace{-0.5cm}
\begin{center}
\begin{tabular}{cccc}
\hline
\textbf{Downstream Task}&\textbf{Task Type}&\textbf{Input}&\textbf{Output}\\
\hline
Direct Positioning&Regression&WC+$\mathrm{Text_{pos}}$&UE Position\\
LOS/NLOS Identification&Classification&WC+$\mathrm{Text_{los}}$&UE LOS Status\\
MIMO Precoding&Regression&WC+$\mathrm{Text_{pre}}$&Precoding Matrix\\
\hline
Beam Selection&Classification&PE+$\mathrm{Text_{beam}}$&Beam Index\\
Path Loss Prediction&Regression&PE+$\mathrm{Text_{pl}}$&Path Loss Value\\
\hline
\multicolumn{4}{c}{\scriptsize (WC and PE denote wireless channel data and physical environment data, respectively.)}
\end{tabular}
\label{t1}
\end{center}
\vspace{-0.7cm}
\end{table}

\begin{figure}[t]
    \centering
    \subfigure[Direct Positioning-W]{
		\begin{minipage}[b]{0.241\textwidth}
	    \centering
		\includegraphics[width=1\textwidth]{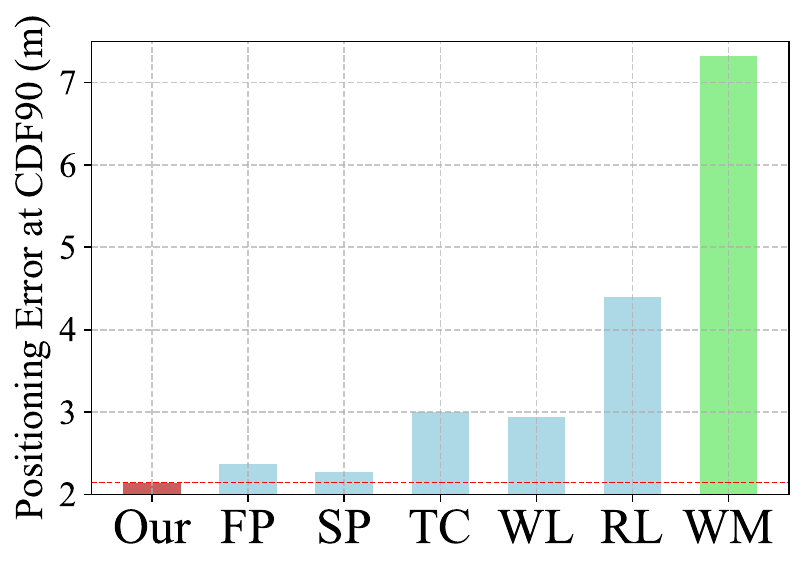} 
        \end{minipage}
    }
    \hspace{-0.6cm}
    \subfigure[Direct Positioning-D]{
		\begin{minipage}[b]{0.241\textwidth}
	    \centering
		\includegraphics[width=1\textwidth]{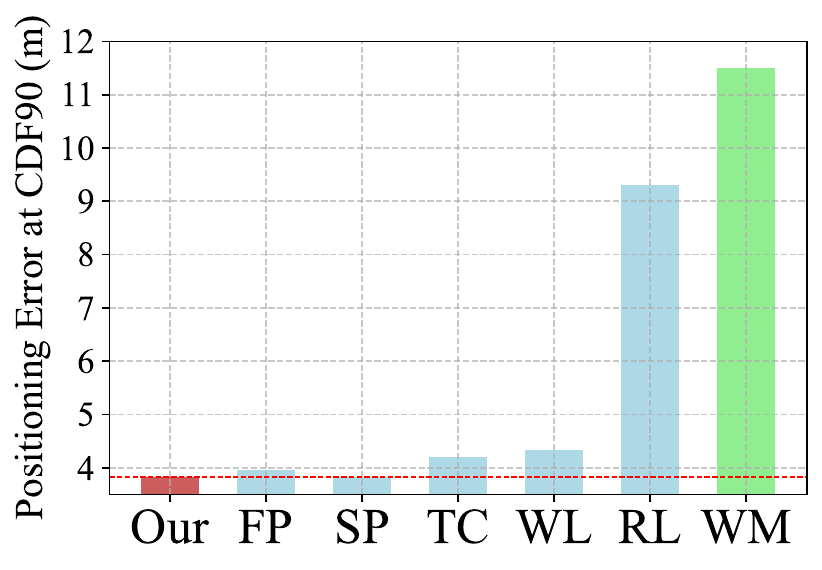} 
        \end{minipage}
    }
    \\
    \vspace{-0.1cm}
	\subfigure[LOS/NLOS Identification-W]{
		\begin{minipage}[b]{0.241\textwidth}
		\centering
   	 	\includegraphics[width=1\textwidth]{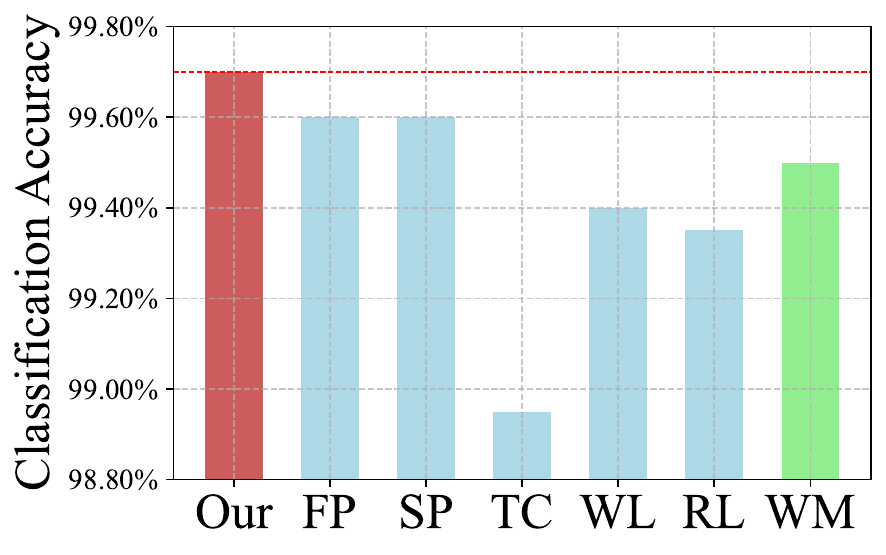}
		\end{minipage}
	}
    \hspace{-0.6cm}
    \subfigure[LOS/NLOS Identification-D]{
		\begin{minipage}[b]{0.241\textwidth}
		\centering
   	 	\includegraphics[width=1\textwidth]{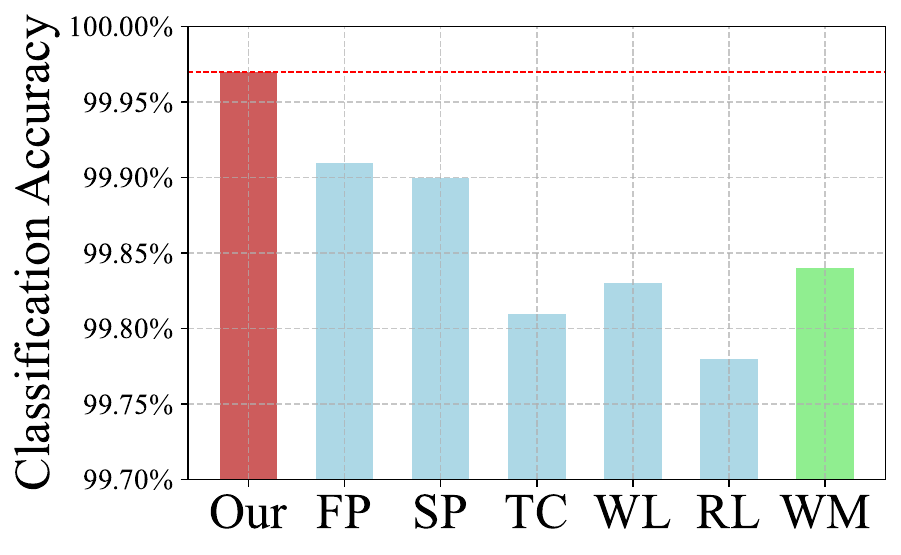}
		\end{minipage}
	}
    \\
    \vspace{-0.1cm}
	\subfigure[MIMO Precoding-W]{
		\begin{minipage}[b]{0.241\textwidth}
		\centering
   	 	\includegraphics[width=1\textwidth]{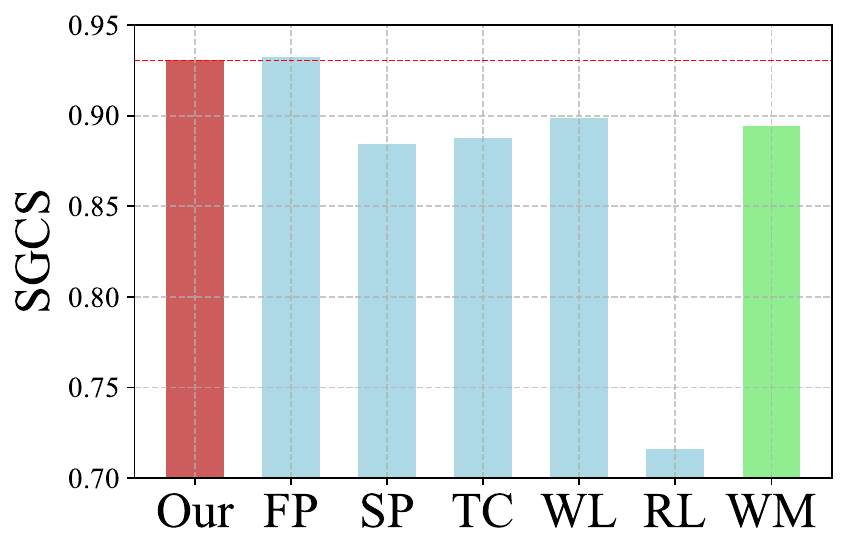}
		\end{minipage}
	}
    \hspace{-0.6cm}
    \subfigure[MIMO Precoding-D]{
		\begin{minipage}[b]{0.241\textwidth}
		\centering
   	 	\includegraphics[width=1\textwidth]{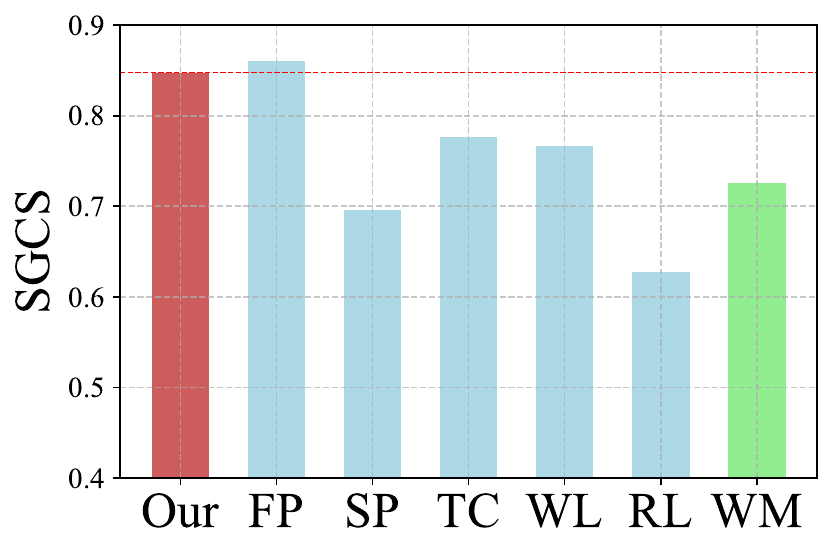}
		\end{minipage}
	}
	\caption{The performance of our proposed method and six benchmarks across the channel-based direct positioning, LOS/NLOS identification, and MIMO precoding tasks. Left: WAIR-D area \#00032. Right: DeepMIMO O1 BS\#12.}
	\label{f4}
\vspace{-0.2cm}
\end{figure}

\begin{figure}[t]
    \centering
    \subfigure[Beam Selection]{
		\begin{minipage}[b]{0.248\textwidth}
	    \centering
		\includegraphics[width=1\textwidth]{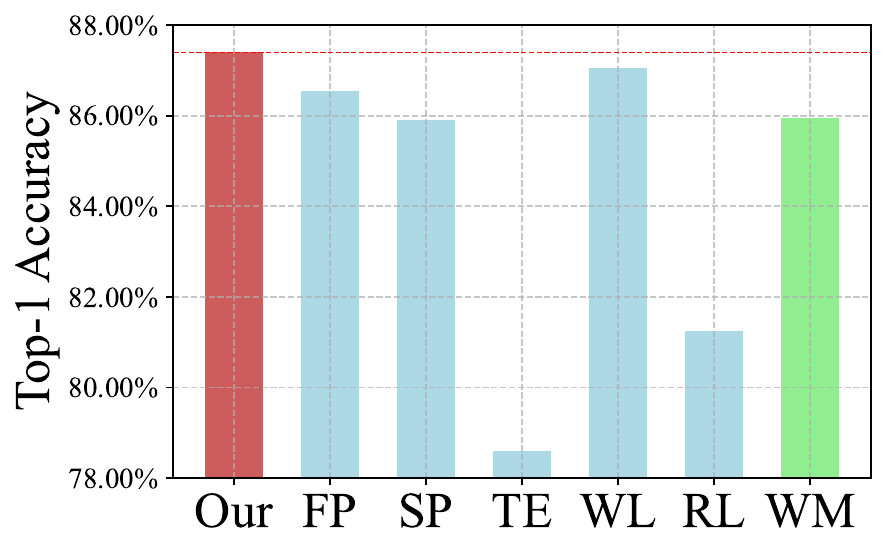} 
        \end{minipage}
    }
    \hspace{-0.6cm}
    \subfigure[Path Loss Prediction]{
		\begin{minipage}[b]{0.234\textwidth}
	    \centering
		\includegraphics[width=1\textwidth]{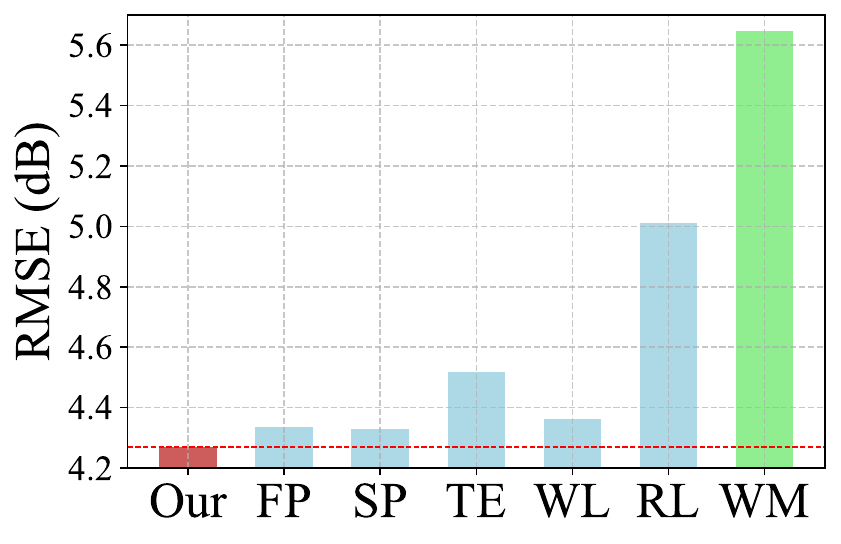} 
        \end{minipage}
    }
	\caption{The performance of our proposed method and six benchmarks across the environment-based beam selection and path loss prediction tasks in WAIR-D area \#00247.}
	\label{f5}
\vspace{-0.5cm}
\end{figure}

\vspace{-0.3cm}
\subsection{Downstream Tasks and Benchmarks}
Five typical use cases, direct positioning, LOS/NLOS identification, MIMO precoding, beam selection, and path loss prediction, are discussed herein, each with distinct dimensionality, precision requirements and tailored optimization strategies. High-precision, channel-based direct positioning is evolving many applications such as mobile navigation and autonomous driving. LOS/NLOS identification, crucial for handover decisions in dynamic environments, directly impacts signal strength and network reliability. MIMO precoding and beam selection mitigate severe path loss at extremely high frequencies, ensuring robust signal transmission. Accurate path loss prediction is essential for estimating transmitter coverage and optimizing wireless network performance. The task types, inputs, and outputs for these use cases are summarized in Table \ref{t1}. Input text includes task keywords such as “position”, “LOS status”, “precoding”, “beam selection" or “path loss" along with task-specific learnable prefix prompts. The loss functions used for the five tasks are mean squared error (MSE) loss, cross-entropy loss, squared generalized cosine similarity (SGCS) loss, focal loss, and MSE loss, respectively. For the precoding task, the CSI undergoes singular value decomposition (SVD) to obtain the optimal precoding matrix. For the beam selection task, the best beam index is chosen from a discrete fourier transform (DFT) codebook.

Considering that we are the first to implement a detailed wireless AI$^2$MMUM, ablation studies are conducted to evaluate the necessity of each module's innovative design. Six benchmarks are used for comparison. The fixed prompt (FP) method utilizes only fixed task key descriptions without learnable prefix prompts as text input, emphasizing the contribution of the latter. The same prompt (SP) method uses a single, identical instruction comprising both fixed (“user information”) and learnable prompts to perform all tasks, highlighting the importance of distinct task instructions for guiding task-related feature extraction. The train EPNN/CFENN (TE/TC) method involves training EPNN or CFENN from scratch, underscoring the feature extraction capability of them derived from large-scale multi-modal alignment. In the without LoRA (WL) method, the LLM backbone processes wireless data solely using its original pre-trained language knowledge, demonstrating the role of LoRA in learning domain-specific knowledge. The random LLM (RL) method uses a randomly initialized and frozen LLM backbone trained with LoRA, assessing whether language knowledge benefits communication task execution. Most importantly, the without LLM (WM) method excludes the Task Instruction Module and LLM backbone, directly connecting the adapter to task heads for end-to-end supervised training, serving as a traditional wireless AI method.

\vspace{-0.2cm}
\subsection{Results and Analysis}
The positioning error corresponding to 90\% of the samples on the cumulative distribution function (CDF) curve of errors is denoted as CDF90. The positioning error at CDF90, classification accuracy, SGCS, top-1 accuracy, and root mean squared error (RMSE) are used as evaluation metrics for direct positioning, LOS/NLOS identification, MIMO precoding, beam selection, and path loss prediction tasks, respectively. The performance of the proposed AI$^2$MMUM, along with six benchmarks, is evaluated across these environment- or channel-based tasks on the WAIR-D and DeepMIMO datasets, as illustrated in Figs. \ref{f4} and \ref{f5}. In summary, our method generally outperforms all benchmarks across all modalities, datasets, and tasks, demonstrating its superior overall performance.

The results of the FP method highlight that incorporating learnable prompts enables the model to more accurately identify and extract task-related features, thereby improving overall performance. In the SP method, the LLM backbone outputs task-agnostic features because all tasks share the same instruction, which may suffice for low-dimensional targets such as position, LOS status, or path loss. However, for high-dimensional targets like the precoding matrix or beam index, these features are influenced by labels from other tasks, introducing biases that significantly degrade prediction accuracy. The TE/TC method underscores the limitations of relying solely on local knowledge, highlighting the robust representation capabilities of EPNN and CFENN achieved through large-scale multi-modal alignment. The performance of the WL method remains intuitively acceptable due to the reduced tuning parameters by excluding LoRA, indicating that the original pre-trained weights of the LLM backbone possess strong convergence properties for communication tasks. LoRA further enhances task performance by effectively absorbing new wireless knowledge. In the RL method, random initialization of the telecom LLM leads to significant performance degradation in most tasks, suggesting that the language knowledge is compatible with wireless domain knowledge. At the same time, LoRA partially compensates for deficiencies caused by initialization. Most importantly, the traditional WM method fails to leverage the LLM backbone's generalization capability and the discriminative power of task instructions, struggling to adapt to multiple tasks with high precision.

\section{Conclusion}
In this paper, we propose the AI$^2$MMUM framework, which leverages the generalization capability of LLM and the discriminative power of task instructions to process 6G-oriented multi-modal data and flexibly perform various downstream tasks. Four key components: a multi-modal radio feature extraction module, a task instruction module, a telecom LLM backbone enhanced with LoRA, and lightweight task-specific heads are specially designed to acquire domain-specific knowledge and improve overall performance. Ablation experiments demonstrate that AI$^2$MMUM consistently outperforms traditional non-LLM methods and models lacking our innovative design in environment/channel-based direct positioning, LOS/NLOS identification, MIMO precoding, beam selection, and path loss prediction tasks. These results highlight the compatibility between radio and language knowledge, heralding a promising future for unified wireless multi-modal intelligence.


\begin{thebibliography}{1}
\bibliographystyle{IEEEtran}
\bibitem{r0} Z. Chen, Z. Zhang, and Z. Yang, ``Big AI models for 6G wireless networks: Opportunities, challenges, and research directions," in \textit{IEEE Wireless Communications}, vol. 31, no. 5, pp. 164-172, October 2024.
\bibitem{r1} L. Bariah, Q. Zhao, H. Zou, Y. Tian, F. Bader, and M. Debbah, ``Large generative AI models for telecom: The next big thing?," in \textit{IEEE Communications Magazine}, vol. 62, no. 11, pp. 84-90, November 2024.
\bibitem{r2} S. Xu, C. K. Thomas, O. Hashash, N. Muralidhar, W. Saad, and N. Ramakrishnan, ``Large multi-modal models (LMMs) as universal foundation models for AI-native wireless systems," in \textit{IEEE Network}, vol. 38, no. 5, pp. 10-20, Sept. 2024.
\bibitem{r3} X. Cao et al., ``MAPLM: A real-world large-scale vision-language benchmark for map and traffic scene understanding," in \textit{Proceedings of the IEEE/CVF conference on computer vision and pattern recognition}, 2024, pp. 21819-21830.
\bibitem{r4} R. Guan et al., ``Talk2Radar: Bridging natural language with 4D mmWave radar for 3D referring expression comprehension," 2024, \textit{arXiv:2405.12821}.
\bibitem{r5} D. Wu et al., ``NetLLM: Adapting large language models for networking," in \textit{Proceedings of the ACM SIGCOMM 2024 Conference}, 2024, pp. 661-678.
\bibitem{r6} T. Jiao et al., ``6G-oriented CSI-based multi-modal pre-training and downstream task adaptation paradigm," in \textit{2024 IEEE International Conference on Communications Workshops (ICC Workshops)}, Denver, CO, USA, 2024, pp. 1389-1394.
\bibitem{r6.5} T. Jiao et al., ``Addressing the curse of scenario and task generalization in AI-6G: A multi-modal paradigm," in \textit{IEEE Transactions on Wireless Communications}, early access, 2025.
\bibitem{r7} Y. Huangfu et al., ``WAIR-D: Wireless AI research dataset," 2022, \textit{arXiv:2212.02159}.
\bibitem{r7.5} A. Alkhateeb, ``DeepMIMO: A generic deep learning dataset for millimeter wave and massive MIMO applications,” 2019, \textit{arXiv:1902.06435}.
\bibitem{r8} Z. Xiao et al., ``LLM agents as 6G orchestrator: A paradigm for task-oriented physical-layer automation," in \textit{2024 IEEE Globecom Workshops (GC Wkshps)}, Cape Town, South Africa, 2024.
\end{thebibliography}
\end{document}